\documentclass[conference]{IEEEtran}
\IEEEoverridecommandlockouts

\usepackage{cite}
\usepackage{amsmath,amssymb,amsfonts}
\usepackage{graphicx}
\usepackage{textcomp}
\usepackage{xcolor}
\usepackage{booktabs}
\usepackage{array}
\usepackage{xurl}
\usepackage[hidelinks]{hyperref}

\newcommand{\cado}{CaDO}
\newcommand{\search}{CaDO-Search}
\newcommand{\ci}[2]{[$#1$, $#2$]}
\DeclareMathOperator*{\argmax}{arg\,max}
\DeclareMathOperator{\Eval}{Eval}
\newcommand{\vd}[1]{\textsf{#1}}

\begin{document}
\bstctlcite{IEEEexample:BSTcontrol}

\title{CaDO: Formal Verification-Guided Multi-Agent Framework for Causal Dataset Repair and Generation}

\author{\IEEEauthorblockN{1\textsuperscript{st} Songli Yu}
\IEEEauthorblockA{\textit{Shandong University} \\
Qingdao, China \\
syu@mail.sdu.edu.cn}
\and
\IEEEauthorblockN{2\textsuperscript{nd} Qi Luo}
\IEEEauthorblockA{\textit{Shandong University} \\
Qingdao, China \\
qiluo@sdu.edu.cn}
\thanks{Corresponding author: Qi Luo.}
}

\maketitle

\begin{abstract}
High-quality causal reasoning datasets are essential for developing models with reliable causal understanding, particularly in specialized domains.
However, existing datasets frequently suffer from incorrect annotations and structural imbalances, which can lead to biased predictions and limited generalization. To address this issue, we propose \cado{} (Causal-Engine-Guided Multi-Agent Data Optimization), a multi-agent framework for automated diagnosis and repair of causal reasoning datasets. Our framework consists of three key stages: formal causal reasoning, multi-level diagnosis, and automated data repair and generation. First, a Formalizer Agent translates natural language questions into formal causal queries to minimize semantic deviations and make retrieval and verification more explicit and tractable. Second, a dual-level diagnosis mechanism employs independent Critic Agents to identify sample-level errors while analyzing query-type coverage and label distributions to detect dataset-level structural imbalances. Finally, the framework employs reusable repair operators grounded in formal causal semantics to correct erroneous samples, while Explorer and Expander Agent collaboratively produce diverse and semantically consistent new samples. Experimental results show that \cado{} significantly outperforms the strongest of five data cleaning baselines in 16 out of 18 experimental settings, improving downstream model accuracy by up to 9.1\%. Moreover, at least 95\% of the samples flagged by \cado{} are defective, and on CLadder, \cado{} incorrectly modifies 92.2\% fewer correct labels than an LLM-based judge, demonstrating its effectiveness and reliability. Our code and data are publicly available at \url{https://github.com/thisissyu/cado}.
\end{abstract}


\section{Introduction}
\label{sec:intro}

\begin{figure}[t]
\centering
\includegraphics[width=\linewidth]{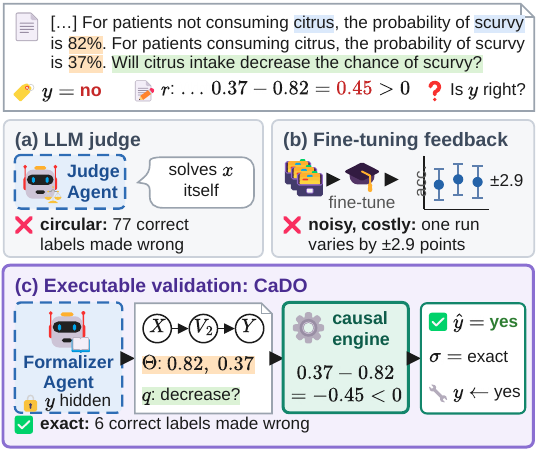}
\caption{Three ways to validate a repair of a causal training example. The CLadder example (abridged) has a flipped label and a rationale that argues for it. (a)~An LLM judge (Judge Agent) decides with the very skill the data should teach, so its errors enter the data. (b)~Fine-tuning feedback scores a whole pipeline, and one run varies more than good pipelines differ (schematic plot). (c)~\cado{} reads the text into a formal causal query without seeing the label, computes the answer with a causal engine, and repairs the label only on firm ground. Unlike (a), no LLM decides the answer, and unlike (b), each example is checked exactly before any model is trained. Numbers from the corrupted CLadder training set: correct labels made wrong by GPT-Judge and by \cado{}, and the standard deviation of one proxy fine-tuning run (Qwen3-4B).}
\label{fig:teaser}
\end{figure}

\begin{figure*}[t]
\centering
\includegraphics[width=\textwidth]{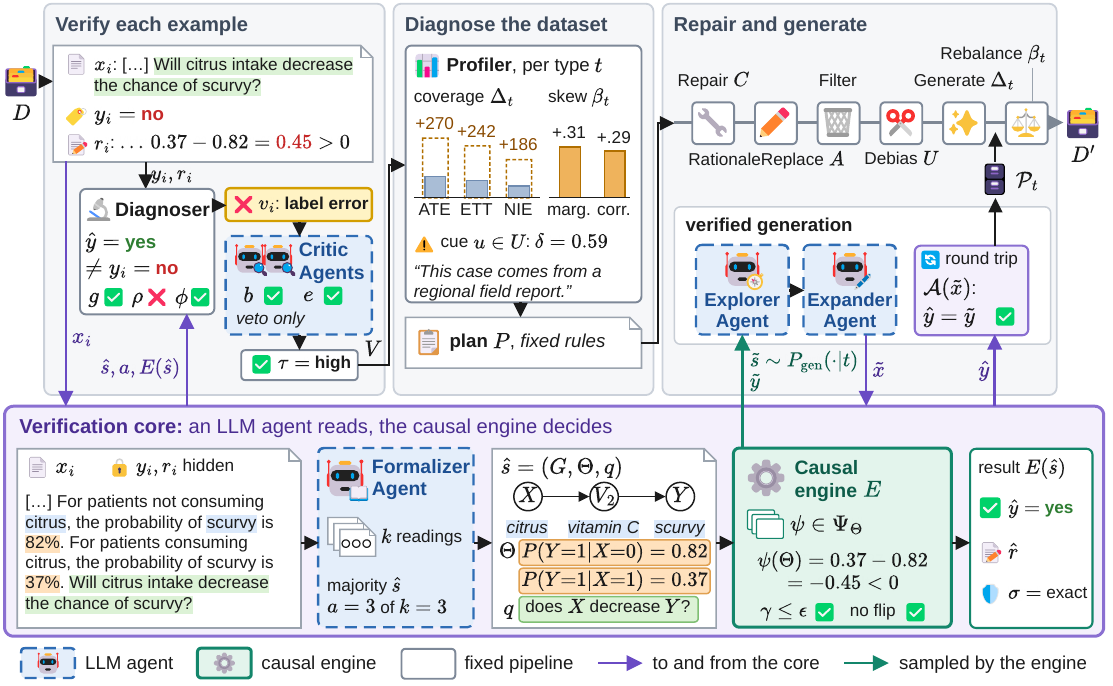}
\caption{Overview of the CaDO framework. CaDO integrates formal causal verification, multi-level diagnosis, and automated dataset repair and generation. \textbf{Bottom:} The verification core uses a label-blind Formalizer Agent to translate natural-language questions into formal causal queries, which are evaluated by a causal engine to produce verified answers, derivations, and verification statuses. \textbf{Top left:} At the sample level, the Diagnoser identifies annotation errors through grounding, rationale consistency, and verification confidence checks, while independent Critic Agents assess the reliability of proposed repairs. \textbf{Top middle:} At the dataset level, the Profiler detects insufficient query-type coverage, label imbalance, and spurious shortcut cues to construct a rule-based repair plan. \textbf{Top right:} Six operators perform targeted correction, filtering, debiasing, and data augmentation. Newly generated examples are validated through a round-trip verification process before inclusion in the repaired dataset. This design ensures that repair and generation decisions are grounded in executable causal semantics rather than LLM judgments.}
\label{fig:main}
\end{figure*}

Large language models (LLMs) are increasingly trained and evaluated on formal causal reasoning, such as estimating the effect of an intervention, answering a counterfactual question, or deciding which causal structures a set of observed dependences allows~\cite{jin2023cladder,jin2024corr2cause,dai2026unico}. The answer to such a question follows from a causal model by the rules of causal inference, so the widely used datasets for this skill are generated by programs rather than collected. CLadder~\cite{jin2023cladder}, for example, samples a causal graph, its probabilities and a query, computes the answer with a causal engine (a program that applies causal inference rules to a formal causal model and query), and writes the question as a short story with everyday variable names and a step-by-step solution. Corr2Cause~\cite{jin2024corr2cause} and CauGym~\cite{chen2026caugym} are built in a similar way.

Generated data can be noisy or incorrect. Even when a program computes the original answer, the text, the rationale and the composition of a dataset come from templates, sampling choices and LLMs, and defects can enter at each step. In CLadder, 2,766 of the 8,532 rationales have an arithmetic line that is off by more than 0.05, and audits report wrong, under-specified and shortcut-prone items in these benchmarks~\cite{dai2026unico,li2025cdcr,clarke2026procedural}. Defects affect single examples, as wrong labels, premises, questions or rationales, and the dataset as a whole, as missing query types, skewed labels or sentences that predict the answer without causal information. A model fine-tuned on such data learns the defects along with the skill, since neural networks memorise wrong labels~\cite{song2023learning} and generator artefacts let models answer from surface patterns~\cite{clarke2026procedural,yu2026caliper}. Today such defects are found by manual audits or by LLM flagging that people check by hand~\cite{dai2026unico,li2025cdcr}, which does not scale, and generating the data again would repeat the same kinds of defects. Our objective is to repair such datasets automatically, in single examples and as a whole, without relying on an LLM's opinion of what is correct.

Existing approaches face three key challenges in causal dataset repairing. (1) LLM-as-a-Judge methods often require substantial token consumption and repeated inference to assess data correctness, resulting in high computational costs and low efficiency. Alternatively, fine-tuning models for error detection may introduce additional noise and propagate existing annotation errors, compromising diagnostic reliability. (2) Causal reasoning datasets may exhibit structural imbalances, such as skewed label distributions and uneven coverage of causal query types. These biases can encourage downstream models to exploit statistical shortcuts rather than learn genuine causal relationships, leading to biased predictions and limited generalization. (3) Repairing causal reasoning datasets requires jointly considering multiple factors, including label correctness, causal consistency, and dataset-level distributional properties. Existing repair processes often rely on extensive manual intervention to identify errors, determine appropriate corrections, and maintain overall dataset quality, making large-scale automated repair challenging.

To address these challenges, we propose \cado{} (Causal-Engine-Guided Multi-Agent Data Optimization), a multi-agent framework that leverages the underlying formal semantics of causal reasoning datasets for automated diagnosis and repair. As illustrated in Fig.~\ref{fig:main}, CaDO consists of three key stages: formal causal reasoning, multi-level diagnosis, and automated data repair and generation. First, we employ a Formalizer Agent to translate natural language questions into formal queries and independently derive their answers, enabling reliable verification of existing labels without relying on the original annotations while reducing semantic deviations. Second, we introduce a dual-level diagnosis mechanism to identify potential defects at both the sample and dataset levels. At the sample level, we employ two independent Critic Agents to verify formal interpretations and mitigate potential interpretation errors. At the dataset level, we analyze query-type coverage and label distributions to identify structural imbalances and potential shortcut cues. Finally, we perform targeted data repair and generation based on the diagnostic results. For sample-level errors, we design reusable repair operators that leverage verifiable formal causal semantics to correct erroneous samples to avoid interference from LLM-induced semantic misjudgments. For dataset-level deficiencies, we leverage Explorer and Expander Agents to generate diverse and semantically consistent new samples while ensuring their correctness, thereby improving query-type coverage and balancing dataset distributions. Together, these components enable CaDO to systematically improve the correctness and structural diversity of causal reasoning datasets while maintaining semantic consistency.

In summary, we make three contributions.
\begin{itemize}
\item We identify formal semantics as the property that makes the correctness of causal training data computable, and we develop executable validation on it. Unlike prior work that runs causal engines in the forward direction, it runs a causal engine backward, from an existing example to its formal query. To our knowledge, we are the first to use a causal engine to repair existing causal reasoning datasets.
\item We propose \cado{}, a multi-agent framework for diagnosing, repairing and completing causal reasoning datasets. In contrast to methods that rely on model judgments, statistical signals or downstream feedback, we ground repair and generation in formal causal semantics, thereby reducing errors in individual examples and improving dataset coverage and label balance.
\item We build corrupt-then-repair test beds on three datasets whose ground truth is hidden from all methods, so that both the repaired data and the models trained on it can be scored exactly, and we evaluate under a pre-registered protocol. \cado{}'s output trains significantly more accurate external evaluation models than the strongest of five baselines in 16 of 18 settings, by up to 9.1 points.
\end{itemize}

\section{Problem Definition}
\label{sec:problem}

This section defines the problem that \cado{} solves by describing the datasets we target, the defects we consider, and the repair task with its measures. Because the correct label of every example in these datasets follows from its text, defects can be defined exactly and a repair can be scored against a hidden ground truth.

\subsection{Causal Datasets with Formal Semantics}

A dataset $D=\{(x_i,y_i,r_i)\}_{i=1}^{n}$ consists of examples with a text $x_i$, which states premises and a question, a label $y_i$ and, optionally, a rationale $r_i$ that derives the label. The dataset has \emph{formal semantics} if for every well-formed example there is a causal model $M_i$ and a query $q_i$ such that the correct label is
\begin{equation}
y_i^{*}=\Eval(M_i,q_i),
\label{eq:eval}
\end{equation}
and $x_i$ is a natural-language rendering of $(M_i,q_i)$. In CLadder and CauGym, $M_i$ consists of a directed acyclic graph (DAG) over binary variables and the quantities that the text states, which are probabilities, conditional probabilities or structural equations. The query $q_i$ is associational, interventional or counterfactual, for example the average treatment effect, the natural indirect effect or the probability of necessity. In Corr2Cause, the text states which pairs of variables are correlated and which independences and conditional independences hold, and $q_i$ asks whether a causal hypothesis holds in every graph consistent with them. We write $t(q)\in T$ for the type of a query $q$.

\subsection{Defects at Two Levels}

An example is defective if its label is wrong, $y_i\neq y_i^{*}$, if its rationale $r_i$ does not derive $y_i^{*}$, or if its text $x_i$ determines no answer at all. At the \emph{example level} we consider five defects. A wrong label comes with a rationale that argues for it. A changed premise alters a stated number or causal link so that the label no longer follows, and a changed question flips the asked direction while the label is kept. A wrong rationale fails to derive the label, and a weak teacher's solution is a wrong solution written by a weaker model.

A dataset can teach badly even when every example is right. Let $p_t=|\{i\mid t(q_i)=t\}|/n$ be the share of query type $t$ and $\pi_t$ its share in the target distribution. At the \emph{dataset level} we consider \emph{coverage gaps}, where $p_t\ll\pi_t$ for some types, \emph{label skew}, where the labels within a type are far from balanced, and \emph{shortcut cues}, sentences that co-occur with one label in training but carry no causal information. We leave out defects that do not change what an external evaluation model learns about the answer, such as differences in difficulty between training and test data, overlap between them, and typos.

\subsection{Task and Measures}

\textbf{Task.} The repair task turns a dataset $D$ with unknown defects into a repaired dataset $D'$ on which an external evaluation model, an LLM fine-tuned on $D'$, answers held-out causal questions as accurately as possible. A method also receives the query-type mix $\pi=(\pi_t)_{t\in T}$ of an unlabelled reference set, in our experiments that of the dev split. The method may change labels and rationales, remove examples and add new ones. It never sees which examples are defective, the defect types, the dataset's generation metadata, such as CLadder's hidden graph and query fields, or any test data. It may call an LLM at a cost, but its decisions about correctness must not rest on an LLM's judgement.

\textbf{Measures.} The main measure is the test accuracy of external evaluation models fine-tuned on $D'$. Because our test beds inject defects into clean data, we know the ground truth of every training example, while all methods are blind to it. We therefore also score the data itself, by the precision and recall of the examples a method flags, the share of wrong labels among the original examples it keeps, the number of defective examples it fixes, and the numbers of correct labels it changes to wrong ones and of correct examples it removes.

\section{The \cado{} Framework}
\label{sec:method}

We design \cado{} to repair and supplement datasets through three parts (Fig.~\ref{fig:main}). First, we use a label-blind Formalizer Agent to produce formal causal queries and an engine to return answers, derivations and exactness statuses. Then, we use the Diagnoser and Critic Agents to check individual examples, and the Profiler to identify coverage gaps, label skew and shortcut cues by query type. We combine both diagnoses into a repair plan using fixed rules. Finally, we execute the plan with six operators. To generate new examples, we compute answers first, use the Explorer Agent to propose scenarios and variable names, and the Expander Agent to word variable values. We render the text with templates and retain only examples yielding the same answer when read back, filling gaps and balancing labels.

\subsection{The Verification Core}
\label{sec:core}

\textbf{Formal causal queries.} The verification core implements executable validation on formal causal queries. A formal query $s=(G,\Theta,q)$ consists of a DAG $G$ whose variables carry roles, such as the treatment $X$ and the outcome $Y$, the stated quantities $\Theta$ as pairs of a term and a value, and the query $q$ with its type $t(q)$ and its direction, for example whether $X$ increases $Y$ through paths that avoid $V_2$. Unobserved variables and deterministic mechanisms are part of $G$ when the text gives them. For Corr2Cause, $\Theta$ lists the stated correlations and independences, and $q$ is a hypothesis about a parent, an ancestor, a collider or a confounder. We write $\mathcal{S}$ for the set of formal queries.

\textbf{Label-blind Formalizer Agent.} The Formalizer Agent reads only the text of an example, never its label or rationale. It is an LLM $f_\theta$ that maps a text $x$ to a distribution $p_\theta(\cdot\mid x)$ over $\mathcal{S}$. We train it without any annotation of the target dataset, on synthetic pairs that the engine produces ``model first, text second''~\cite{li2025sqlfactory},
\begin{equation}
\begin{aligned}
D_{\mathrm{syn}}&=\{(\mathrm{render}(s),s)\mid s\sim P_{\mathrm{gen}}\},\\
\theta&=\argmax_{\theta'}\textstyle\sum_{(x,s)\in D_{\mathrm{syn}}}\log p_{\theta'}(s\mid x).
\end{aligned}
\label{eq:formalizer}
\end{equation}
The generator $P_{\mathrm{gen}}$ samples a random causal model on a graph family of the target dataset, computes every stated number exactly from the model's joint distribution, and renders the text in several phrasing families with variable names from a bank that an LLM writes. We remove the names that occur in the target dataset from the bank, so the Formalizer Agent cannot memorise the target's stories, and we check every pair by solving $s$ with the engine. At inference time the Formalizer Agent draws $k$ readings $s^{(1)},\dots,s^{(k)}\sim p_\theta(\cdot\mid x)$ at temperature $\tau_{\mathrm{F}}$. A canonical form $c(\cdot)$ aligns the variables by role and sorts the terms. The reading we use is the most frequent canonical form, and its agreement is the number of readings that share it,
\begin{equation}
\begin{aligned}
\hat s(x)&=\argmax_{c}\,\bigl|\{j\mid c(s^{(j)})=c\}\bigr|,\\
a(x)&=\bigl|\{j\mid c(s^{(j)})=\hat s(x)\}\bigr|.
\end{aligned}
\label{eq:majority}
\end{equation}

\textbf{A causal engine built for verification.} For a reading $s$, the engine $E$ returns an answer $\hat y$, a derivation $\hat r$ in the dataset's rationale format, and a status $\sigma$,
\begin{equation}
E(s)=(\hat y,\hat r,\sigma).
\label{eq:engine}
\end{equation}
For CLadder and CauGym, it enumerates the set $\Psi(G,q)$ of identification formulas that are valid for $G$ and $q$. These are back-door adjustment over every admissible set, the front-door and instrumental-variable formulas, the mediation formulas for natural direct and indirect effects, the standard expressions for the effect of treatment on the treated, explaining away and collider bias, deterministic counterfactuals from structural equations, and the bounds of Tian and Pearl for the probabilities of necessity and sufficiency~\cite{pearl2009causality,tian2000probabilities}. For an admissible set $Z$, for instance, back-door adjustment gives
\begin{equation}
P(y\mid \mathrm{do}(x))=\textstyle\sum_{z}P(y\mid x,z)\,P(z).
\label{eq:backdoor}
\end{equation}
Let $\Psi_\Theta\subseteq\Psi(G,q)$ be the formulas whose terms $\Theta$ states, and let $\mathrm{dec}_q$ map a value to an answer, for example to ``yes'' if an effect that the question asks to be positive exceeds 0. The engine evaluates every $\psi\in\Psi_\Theta$, measures how far the formulas disagree, and tests whether the answer could flip within the rounding of the stated numbers,
\begin{equation}
\begin{aligned}
\gamma&=\textstyle\max_{\psi,\psi'\in\Psi_\Theta}|\psi(\Theta)-\psi'(\Theta)|,\\
\mathit{flip}&=\bigl[\mathrm{dec}_q(\psi(\Theta'))\neq\mathrm{dec}_q(\psi(\Theta))\text{ for some }\Theta'\in R(\Theta)\bigr],
\end{aligned}
\label{eq:spread}
\end{equation}
where $R(\Theta)$ is the set of quantities that round to the stated ones. It returns $\hat y=\mathrm{dec}_q(\psi(\Theta))$ with one of five statuses,
\begin{equation}
\sigma=\begin{cases}
\vd{non-identifiable} & \text{if } \Psi(G,q)=\emptyset,\\
\vd{underdetermined} & \text{if } \Psi_\Theta=\emptyset,\\
\vd{conflict} & \text{if } \gamma>\epsilon,\\
\vd{borderline} & \text{if } \mathit{flip},\\
\vd{exact} & \text{otherwise,}
\end{cases}
\label{eq:status}
\end{equation}
with $\epsilon=0.02$. The rules of causal inference it applies are standard. What it adds for verification is the cross-check of all applicable formulas and the status, which tells \cado{} when to abstain.

\textbf{Conventions and Corr2Cause.} Two further features make the engine's output usable as training data. Where a dataset's generator uses a formula that is not exact for the graph, the engine reproduces it as an explicitly flagged \emph{convention}. CLadder, for instance, computes the effect of treatment on the treated in a confounded graph with $P(z)$ instead of $P(z\mid x)$. The engine also writes $\hat r$ in the dataset's own rationale format, so a repaired example looks like a clean one. For Corr2Cause, it follows the dataset's definition~\cite{jin2024corr2cause}. Under the Markov and faithfulness assumptions, it enumerates the Markov equivalence class $[G]_\Theta$ of the DAGs that are consistent with the stated dependences and independences, and for the hypothesis $h$ of $q$ it sets
\begin{equation}
\hat y=\bigl[\,h \text{ holds in every } G'\in[G]_\Theta\,\bigr].
\label{eq:c2c}
\end{equation}
The same engine underlies $P_{\mathrm{gen}}$, so training, verification and generation share one semantics.

\subsection{Verify Each Example}
\label{sec:verify}

\begin{table}[t]
\centering
\caption{Verdicts of the Diagnoser, their conditions and the repair each one proposes. A repair that changes a label requires firm ground $\phi(x)$, eq.~\eqref{eq:firm}.}
\label{tab:verdicts}
\footnotesize
\setlength{\tabcolsep}{3pt}
\begin{tabular}{>{\raggedright\arraybackslash}p{1.75cm}>{\raggedright\arraybackslash}p{3.8cm}>{\raggedright\arraybackslash}p{2.15cm}}
\toprule
Verdict & Condition & Repair \\
\midrule
\vd{consistent} & $\hat y=y$, rationale consistent & none \\
\vd{rationale error} & $\hat y=y$, rationale's numbers, graph, arithmetic or conclusion inconsistent & rationale $\leftarrow\hat r$ \\
\vd{label error} & $\hat y\neq y$ on firm ground, rationale follows the text & label and rationale from the engine \\
\vd{premise mismatch} & $\hat y\neq y$ on firm ground, rationale uses numbers or links the text does not state & label and rationale from the engine \\
\vd{unanswerable} & stable, grounded reading, but the text determines no answer & remove \\
\vd{unresolved} & no reading, a \vd{conflict}, or $\hat y\neq y$ without firm ground & none (abstain) \\
\bottomrule
\end{tabular}
\end{table}

For every example, the Diagnoser decides whether its label and rationale agree with the verification core and whether that decision is firm enough to act on. It runs the core and three checks.

\textbf{Grounding.} A reading is useful only if it describes the text. We require
\begin{equation}
g(x,s)=g_{\mathrm{num}}(x,s)\wedge g_{\mathrm{name}}(x,s)\wedge g_{\mathrm{q}}(x,s),
\label{eq:grounding}
\end{equation}
where $g_{\mathrm{num}}$ holds if the numbers in $x$ are exactly those in $\Theta$, $g_{\mathrm{name}}$ if every variable name of $s$ occurs in $x$, and $g_{\mathrm{q}}$ if the question sentence of $x$ mentions the treatment and the outcome of $q$.

\textbf{Rationale consistency.} A rationale $r$ is consistent with a reading, $\rho(r,s,y)=1$, if its numbers occur in $\Theta$, its graph matches $G$, its last arithmetic line is correct and it concludes $y$.

\textbf{Verdicts.} \cado{} may act on an example only if its reading is stable and the decision rests on firm ground. With $(\hat y,\hat r,\sigma)=E(\hat s(x))$, a reading is \emph{stable} if all $k$ readings agree, and the decision rests on \emph{firm ground} if
\begin{equation}
\phi(x)=[\sigma=\vd{exact}]\wedge[\text{no convention}]\wedge[a(x)=k]\wedge g(x,\hat s(x)).
\label{eq:firm}
\end{equation}
The verdict function assigns each example one of the six verdicts of Table~\ref{tab:verdicts}. If the engine agrees with the label, the example is \vd{consistent} or has a \vd{rationale error}, depending on $\rho$. If the engine disagrees, a change of label is proposed only if $\phi(x)$ holds. The example is then a \vd{premise mismatch} if its rationale uses numbers or links that $\hat s(x)$ does not contain, and a \vd{label error} otherwise. In both cases the label becomes what the text implies, so the distinction only describes the defect. An example whose text determines no answer, with the status \vd{underdetermined} or \vd{non-identifiable}, is \vd{unanswerable} if its reading is stable and grounded. Every other example is \vd{unresolved}, and the Diagnoser abstains on it.

\textbf{Critic Agents and confidence tiers.} Two Critic Agents check the reading of every example that the Diagnoser would change or remove, and they can veto the reading but never supply an answer. The Back-Translation Critic Agent reads $x$ together with $\hat s(x)$ and judges whether the reading is faithful to the text in its links, its numbers and the direction of its question, and we write $b(x)=1$ if it is. The Independent Critic Agent writes its own reading $s'$ of $x$ from a few examples, the engine solves it, and $e(x)=1$ if the result agrees with that of $\hat s(x)$. The Critic Agents and the firm-ground conditions give each proposed repair a confidence tier,
\begin{equation}
\tau(x)=\begin{cases}
\textsf{high} & \text{if } \sigma=\vd{exact}\wedge g\wedge a=k\wedge b\wedge e,\\
\textsf{medium} & \text{if } \sigma=\vd{exact}\wedge g_{\mathrm{num}}\wedge a\geq 2\wedge (b\vee e),\\
\textsf{low} & \text{otherwise,}
\end{cases}
\label{eq:tier}
\end{equation}
with $g$, $a$, $b$ and $e$ evaluated at $x$. \cado{} applies only high-tier repairs, a default we fixed before any test result was seen.

\subsection{Diagnose the Dataset}
\label{sec:diagnose}

\textbf{Profile.} Dataset-level defects show only when the examples are counted per query type. The Profiler therefore groups the examples by the query type of their reading, $t_i=t(\hat s(x_i))$, and computes for every type $t$ the count, the coverage deficit relative to the reference mix, and the label imbalance,
\begin{equation}
\begin{aligned}
n_t&=|\{i\mid t_i=t\}|,\qquad \Delta_t=\max\bigl(0,\,\mathrm{round}(\pi_t n)-n_t\bigr),\\
\beta_t&=\tfrac{1}{n_t}\textstyle\sum_{t_i=t}[y_i=\text{yes}]-\tfrac{1}{2}.
\end{aligned}
\label{eq:profile}
\end{equation}

\textbf{Shortcut scan.} A shortcut cue is a sentence that predicts the label within a type but carries no causal information. The Profiler looks within each type for sentences $u$ that contain no number, are not the question, and occur in at least 20 examples and at least 3\% of the type. It flags $u$ if its presence shifts the type's rate of ``yes'' labels,
\begin{equation}
\delta(u,t)=\bigl|\hat P_t(\text{yes}\mid u\in x)-\hat P_t(\text{yes}\mid u\notin x)\bigr|\geq 0.2,
\label{eq:shortcut}
\end{equation}
where $\hat P_t$ is the empirical rate within type $t$, and $U$ denotes the set of flagged sentences.

\textbf{Plan.} The plan $P$ follows from fixed rules, not from fine-tuning feedback. It repairs the label errors, premise mismatches and rationale errors of the high tier and replaces every verified rationale with the engine's. It removes the unresolved and unanswerable examples and every sentence in $U$. Finally, it generates $\Delta_t$ examples of each type $t$ and rebalances every type to $\beta_t=0$.

\subsection{Repair and Generate}
\label{sec:repair}

\textbf{Operators.} Six operators execute the plan, and each acts only on verified information. Every operator maps a dataset to a dataset, and none rewrites text freely. Let $C$ be the examples with a high-tier repair and $A$ the verified examples, those that are \vd{consistent} or repaired. \emph{Repair} sets $(y_i,r_i)\leftarrow(\hat y_i,\hat r_i)$ for $i\in C$, or only $r_i$ for a rationale error. \emph{RationaleReplace} sets $r_i\leftarrow\hat r_i$ for $i\in A$. \emph{Filter} removes every example whose verdict is \vd{unresolved} or \vd{unanswerable}, and \emph{Debias} deletes every sentence of $U$ from the texts that contain it. \emph{Generate} adds $\Delta_t$ examples of type $t$ from a verified pool $\mathcal{P}_t$, and \emph{Rebalance} adds $|n_t^{\text{yes}}-n_t^{\text{no}}|$ pool examples of the minority label of each type $t$. \cado{} applies the operators once, in this order, with the parameters of the plan,
\begin{equation}
\begin{aligned}
D'=\;&\mathrm{Rebalance}\circ\mathrm{Generate}\circ\mathrm{Debias}\\
&\circ\mathrm{Filter}\circ\mathrm{RationaleReplace}\circ\mathrm{Repair}\,(D).
\end{aligned}
\label{eq:pipeline}
\end{equation}

\textbf{Verified generation.} New examples are built model first. For a query type $t$ and a graph family, the engine samples a formal query $\tilde s\sim P_{\mathrm{gen}}(\cdot\mid t)$ and computes its answer and derivation, $(\tilde y,\tilde r,\tilde\sigma)=E(\tilde s)$. Two generation agents supply the language. The Explorer Agent proposes a realistic everyday scenario for the graph, choosing variable names for its roles that avoid the names already used, and the Expander Agent writes the phrases for each variable's values, such as ``smokers'' and ``nonsmokers''. Templates in the dataset's format then render the text $\tilde x$. A candidate enters the pool only if it survives a round trip through the verification core,
\begin{equation}
\mathcal{A}(\tilde x)=\bigl[E(\hat s(\tilde x)).\hat y=\tilde y\bigr]\wedge g(\tilde x,\hat s(\tilde x))\wedge \neg\,\mathrm{dup}(\tilde x),
\label{eq:roundtrip}
\end{equation}
where $\mathrm{dup}$ excludes near-duplicates of training examples and, to avoid any overlap with test data, candidates that reuse the variable names of a CLadder test story or repeat a Corr2Cause dev or test premise. For CauGym the Explorer Agent proposes the variables together with their value phrases, and for Corr2Cause the templates alone write the text. Every added example therefore carries a label and a derivation computed by the engine, and the Formalizer Agent has read its text back to the same answer.

\section{Experiments}
\label{sec:experiments}

This section evaluates whether executable validation makes repairs precise and the repaired data useful for training. First, we describe the setup and compare the external evaluation models trained on each method's output. Then, we score each method's data against the hidden ground truth, analyze \cado{}'s components and control for the data it adds. Next, we compare its fixed rules with plans chosen by fine-tuning feedback and vary the defect rate and the amount of generated data. Finally, we run \cado{} on the original datasets and measure its cost.

\subsection{Experimental Setup}
\label{sec:setup}

\begin{table}[t]
\centering
\caption{Datasets. ``Wrong'': example-level defects in the corrupted training set, hidden from all methods. The headline test sets were fixed in a pre-registration.}
\label{tab:data}
\footnotesize
\setlength{\tabcolsep}{2pt}
\begin{tabular}{lccc}
\toprule
& CLadder & CauGym & Corr2Cause \\
\midrule
Answer & yes/no & value, yes/no & valid or not \\
Query types & 10 & 7 & 6 \\
Clean train & 6,963 & 3,149 & 6,001 \\
Corrupted (wrong) & 3,599 (1,080) & 1,863 (554) & 3,886 (1,166) \\
Dev & 997 & 350 & 1,076 \\
Headline test & 1,989 & 1,400 & 1,162 \\
Secondary tests & dev, CounterBench & 3 official sets & dev, 2 official sets \\
Metric & accuracy & official (value) & accuracy \\
\bottomrule
\end{tabular}
\end{table}

Our test beds are corrupted copies of three causal reasoning datasets, into which we inject seven types of defects. We know the ground truth of every training example, but no method sees it, so we can score both the external evaluation models trained on a method's output and the output itself. Each output trains six external evaluation models from three families, which gives 18 settings of dataset and model.

\textbf{Datasets.} The three datasets differ in their answers and query types, as Table~\ref{tab:data} shows. CLadder~\cite{jin2023cladder}, in version 1.5, asks yes/no questions of ten query types on all three rungs of the ladder of causation, over ten graphs and 47 stories. We split it by graph and story into train, dev and test, so that no story crosses splits. CauGym~\cite{chen2026caugym} has 3,500 training problems of seven query types on ten-variable graphs. Its answers are probabilities with four decimals, or yes/no for effect questions, and its official metric requires the final value to be within 0.0005 of the truth. Corr2Cause~\cite{jin2024corr2cause} asks whether a causal hypothesis follows from stated correlations and independences. We use a stratified sample of 6,001 rows of its official training set, without the rows whose premise occurs in the dev or test set, and its official dev and test sets.

\textbf{Clean copies.} The ground truth comes from a clean copy of each training set, built with the dataset's own metadata, which the test bed may use but no method ever sees. We recompute every label with the engine and drop the few rows whose label contradicts their text or cannot be determined, 163 of 7,126 on CLadder and 1 on CauGym. The training target of every example is a derivation followed by a final-answer line. For CLadder and Corr2Cause the derivation is the engine's, and for CauGym it is CauGym's own formal steps. Dev and test sets are never modified.

\textbf{Injected defects.} We inject the same seven defect types into the training split of every dataset, at the same rates. At the example level, 30\% of the rows are defective, 6\% for each of the five types. These defects flip the label together with the rationale's conclusion, change one stated number so that the true answer flips, remove or reverse one causal link, flip the asked direction while keeping the label, or replace the solution and answer with those of a weak teacher, the untuned Qwen3-4B. For Corr2Cause, the changed number is one pair's independence statements. At the dataset level, each defect affects three query types of CLadder and two of CauGym and Corr2Cause. A coverage gap keeps only 15\% of the rows of these types, label skew raises the share of one label in them to 85\%, and a neutral cue sentence is added to 80\% of their ``yes'' rows and 20\% of their ``no'' rows. Each defect is mapped to the dataset's format, and a changed premise or question is kept only if the engine confirms that the true answer changed. We set the rates on CLadder, where this setting gave the largest gap between clean and corrupted data on dev among four pre-specified settings, and reused them with two adaptations fixed before any method was run. CauGym, whose gap was small, keeps 5\% instead of 15\% of the under-covered types, chosen by a pre-specified rule. Corr2Cause, whose labels are 85\% ``no'', is skewed towards its majority label, to 98\%. Because coverage gaps and label skew remove rows, the corrupted sets are smaller than the clean ones.

\textbf{Headline test sets.} Before any comparison of methods, we pre-registered one headline test set per dataset. CauGym's official base, rephrased and omitted test sets use a shorter surface format than its training problems and vary strongly across training seeds, so its headline is the pooled deconfounding and redundant sets, which match the training format. A pre-registered check confirmed that clean data beats corrupted data on this headline by more than 3 points with Qwen3-4B, by +4.2 \ci{+2.1}{+6.3}. All other test sets are secondary.

\textbf{External evaluation models.} We fine-tune six instruction-tuned LLMs from three families on each method's output, Qwen3-4B and Qwen3-8B~\cite{yang2025qwen3}, Llama-3.2-3B and Llama-3.1-8B~\cite{grattafiori2024llama3}, and Ministral-3-3B and Ministral-3-8B~\cite{liu2026ministral3}. Each is trained with LoRA~\cite{hu2022lora} of rank 8 and $\alpha=32$ on all linear layers, with a learning rate of $2\cdot10^{-4}$ and cosine decay, an effective batch size of 8, the loss on the answer tokens only, and the reasoning mode disabled. The number of epochs, from 1 to 3, is chosen per dataset and model on the clean dev split, with one rule for all methods. We decode greedily with vLLM~\cite{kwon2023vllm}. The answer is read from the last final-answer line, and a malformed or truncated output counts as wrong. For \cado{}'s Formalizer Agent we fine-tune Qwen3-4B, which read real dev text more accurately than Qwen3-8B, with LoRA on 12,000 CLadder-style, 8,000 CauGym-style or 8,000 Corr2Cause-style synthetic pairs, one Formalizer Agent per dataset, and draw $k=3$ readings at temperature 0.7. All runs, about 960 fine-tuning runs in total, used NVIDIA RTX~3090 GPUs.

\textbf{Compared methods.} Two reference arms and five cleaning baselines receive the same corrupted training set, and every output trains the same external evaluation models with the same recipe. \emph{Corrupted} uses the data as is, and \emph{Clean} uses the clean copy, an oracle rather than a competitor. \emph{Loss filter} removes the 20\% of rows with the highest loss under a Qwen3-4B trained on the data. \emph{Confident learning}~\cite{northcutt2021confident} prunes the rows whose label the data contradicts, with out-of-sample probabilities from four-fold cross-fitting of Qwen3-4B. \emph{GPT-Agree} asks an LLM to answer each question without the label and drops the rows where it disagrees. \emph{GPT-Judge} shows an LLM the question, rationale and label and lets it keep the row, fix it with a new label and rationale, or drop it. \emph{LLM-AutoDP}~\cite{huang2026llmautodp} combines a cleaning team, an optimization team that rewrites with an LLM, a generation team that writes new examples with an LLM, and a selection team, guided by an agent that sees only proxy scores. We reimplemented it for our datasets, with an LLM quality score in place of the original's trained screening model, rule-based text filters and gradient-based selection, and with the same LLM, proxy and search budget as \search{} below, keeping the strategy with the best proxy score. All LLM calls of \cado{} and of the LLM-based baselines use the same model, OpenAI's GPT-6.1 Sol, with low reasoning effort for per-example calls and medium effort for the planning agents. Using one model for every method keeps the LLM a controlled variable of the comparison.

\textbf{\search{} and Random search.} To test whether fine-tuning feedback can choose better plans than \cado{}'s fixed rules, we build \search{}, an internal control that differs from \cado{} only in how its plan is validated. It keeps the verification core, the Diagnoser and the operators, and replaces the fixed planning rules with the outer loop of LLM-AutoDP~\cite{huang2026llmautodp}. In each of up to five rounds, a Planner Agent sees the profile, the operator specifications and the strategies evaluated so far, and proposes four strategies, each an ordered list of operators with parameters. A strategy is scored by fine-tuning a proxy model on two 50\% samples of its output and averaging its accuracy on the verified part of the dev split, and it replaces the current best only if its gain exceeds a noise margin estimated from repeated runs on the raw data. A Reflector Agent summarises what helped, and the search stops after two rounds without improvement. \emph{Random search} draws strategies over \cado{}'s own operators with the same budget and proxy, as a control for the Planner Agent rather than a cleaning baseline. With the Llama and Ministral models, we run the two strongest baselines on the Qwen models, LLM-AutoDP and GPT-Judge, together with Corrupted, Clean and Random search.

\textbf{Protocol.} We pre-registered the statistical protocol. The value of an arm is its mean accuracy over all seeds run. We report differences with 95\% two-level paired bootstrap intervals, which resample each arm's training seeds and then the test items, with 4,000 draws, or 20,000 when a bound is within 0.3 points of 0, so training noise is included. Every arm starts with three seeds. For each pre-specified comparison, an interval that excludes 0 is conclusive. An estimate below 1 point stops the comparison with ``no effect of 2 points or more detected''. Otherwise, both arms receive the same additional seeds, up to eight. Two points is the smallest effect of interest, the effect that eight seeds can resolve given the measured noise. Because test items share stories in CLadder and premises in Corr2Cause, we also report a cluster bootstrap over these groups. All selections, of epochs, of the defect setting and of design variants, use dev data only.

\begin{table*}[t]
\centering
\caption{Test accuracy (\%) of external evaluation models fine-tuned on each version of the corrupted training data (mean over all seeds, headline test sets). Best BL: the most accurate of the five cleaning baselines for that model (A = LLM-AutoDP, J = GPT-Judge, G = GPT-Agree). $^{\dagger}$: \cado{} is significantly better than that baseline (95\% two-level paired bootstrap interval above 0, equal seeds). It is never significantly worse. Clean is the uncorrupted oracle.}
\label{tab:main}
\footnotesize
\setlength{\tabcolsep}{3.2pt}
\begin{tabular}{lrrrrrrrrrrrr}
\toprule
External & \multicolumn{4}{c}{CLadder} & \multicolumn{4}{c}{CauGym} & \multicolumn{4}{c}{Corr2Cause} \\
\cmidrule(lr){2-5}\cmidrule(lr){6-9}\cmidrule(lr){10-13}
evaluation model & Corr. & Best BL & \cado{} & Clean & Corr. & Best BL & \cado{} & Clean & Corr. & Best BL & \cado{} & Clean \\
\midrule
Qwen3-4B & 76.1 & 89.9\textsuperscript{A} & \textbf{94.4}$^{\dagger}$ & 96.7 & 35.6 & 35.8\textsuperscript{G} & \textbf{38.0} & 39.4 & 88.6 & 91.6\textsuperscript{A} & \textbf{95.6}$^{\dagger}$ & 97.1 \\
Qwen3-8B & 75.8 & 87.2\textsuperscript{A} & \textbf{95.2}$^{\dagger}$ & 96.1 & 38.4 & 38.8\textsuperscript{A} & \textbf{39.6} & 41.3 & 88.6 & 92.3\textsuperscript{J} & \textbf{96.9}$^{\dagger}$ & 97.9 \\
Llama-3.2-3B & 84.8 & 86.3\textsuperscript{A} & \textbf{95.4}$^{\dagger}$ & 93.5 & 23.9 & 23.9\textsuperscript{J} & \textbf{27.1}$^{\dagger}$ & 24.5 & 87.8 & 88.1\textsuperscript{A} & \textbf{95.5}$^{\dagger}$ & 97.0 \\
Llama-3.1-8B & 86.2 & 88.5\textsuperscript{A} & \textbf{94.6}$^{\dagger}$ & 95.4 & 29.8 & 30.3\textsuperscript{A} & \textbf{32.3}$^{\dagger}$ & 33.3 & 88.2 & 91.4\textsuperscript{J} & \textbf{97.0}$^{\dagger}$ & 97.7 \\
Ministral-3-3B & 91.3 & 89.4\textsuperscript{A} & \textbf{96.0}$^{\dagger}$ & 97.2 & 31.4 & 31.6\textsuperscript{A} & \textbf{33.2}$^{\dagger}$ & 33.1 & 89.5 & 91.1\textsuperscript{A} & \textbf{96.6}$^{\dagger}$ & 97.6 \\
Ministral-3-8B & 91.1 & 90.9\textsuperscript{A} & \textbf{96.1}$^{\dagger}$ & 97.0 & 39.0 & 37.9\textsuperscript{A} & \textbf{40.0}$^{\dagger}$ & 43.1 & 89.4 & 92.9\textsuperscript{J} & \textbf{97.0}$^{\dagger}$ & 98.1 \\
\bottomrule
\end{tabular}
\end{table*}

\begin{figure*}[t]
\centering
\includegraphics[width=\textwidth]{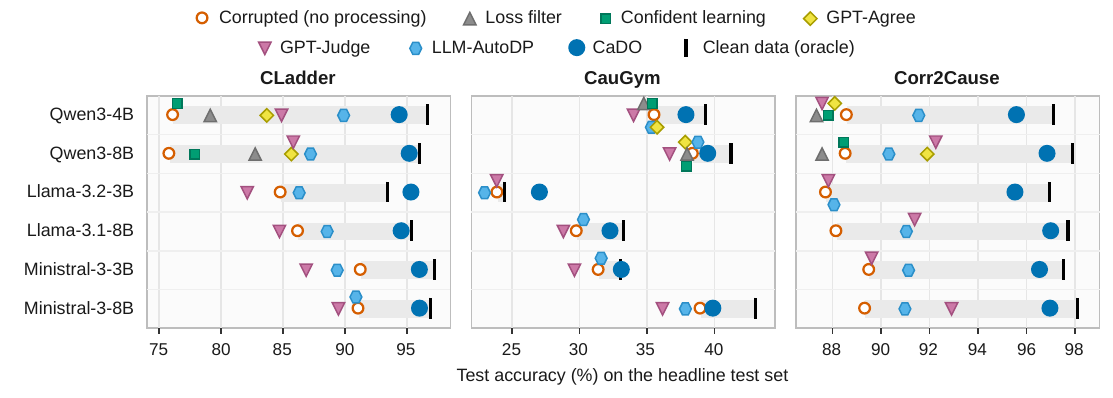}
\caption{Test accuracy of external evaluation models fine-tuned on each method's output (mean over all seeds, headline test sets). Grey band: from the corrupted to the clean training data (oracle). Loss filter, confident learning and GPT-Agree were run with the Qwen models only. Overlapping markers are shifted within their row. Table~\ref{tab:main} gives the values with significance marks.}
\label{fig:mainacc}
\end{figure*}

\subsection{Main Comparison}
\label{sec:main}

\cado{}'s output trains significantly more accurate external evaluation models than the strongest baseline's in 16 of 18 settings, and in no setting is it significantly less accurate than any baseline. Table~\ref{tab:main} compares it with the strongest of the five cleaning baselines for each model, and Fig.~\ref{fig:mainacc} shows every method in every setting. The lead is 4.0 to 9.1 points on CLadder and 4.0 to 7.5 on Corr2Cause. In eleven of these twelve settings \cado{} recovers between 79.9\% and 95.9\% of the accuracy that corruption costs, and with Llama-3.2-3B on CLadder it recovers more than all of it. It stays within 3.5 points of clean data in every setting, and with Llama-3.2-3B it is significantly better than clean data on CLadder, by +1.9 \ci{+0.4}{+3.3}, and on CauGym, by +2.6 \ci{+0.6}{+4.5}.

CauGym is the hard case, because its official metric scores a four-decimal final value that depends mostly on the model's arithmetic. Even clean data beats corrupted data by only 0.6 to 4.1 points. \cado{} is significantly ahead in the four Llama and Ministral settings. With the Qwen models the differences, +0.8 and +0.1, are below one point, and the sequential rule stops with no effect of 2 points or more detected.

The results hold beyond the headline test sets. Against the strongest baseline of the headline test, \cado{} is significantly better for at least five of the six models on the CLadder dev split and on the Corr2Cause dev, paraphrased and refactorized sets, and never significantly worse. On CauGym's official sets, it beats the corrupted data for five models and the strongest baseline for one to three. Eleven of the 12 significant results survive a cluster bootstrap over test stories and premises, all but Qwen3-4B on CLadder against LLM-AutoDP. CounterBench, an out-of-distribution test, barely reacts to data quality, with clean data 1.7 to 5.3 points above corrupted data, and there \cado{} is significantly ahead of the strongest baseline with one model and behind it with another.

\begin{table*}[t]
\centering
\caption{Training data of each method scored against the hidden ground truth (number of defective rows in brackets). Err.: wrong labels among the kept original rows (\%, wrong answer values for CauGym). Fixed: defective rows kept with a corrected label. Dam.: correct labels changed to wrong ones. Rem.: correct rows removed.}
\label{tab:dq}
\footnotesize
\setlength{\tabcolsep}{3.2pt}
\begin{tabular}{lrrrrrrrrrrrr}
\toprule
& \multicolumn{4}{c}{CLadder (1,080)} & \multicolumn{4}{c}{CauGym (554)} & \multicolumn{4}{c}{Corr2Cause (1,166)} \\
\cmidrule(lr){2-5}\cmidrule(lr){6-9}\cmidrule(lr){10-13}
Method & Err. & Fixed & Dam. & Rem. & Err. & Fixed & Dam. & Rem. & Err. & Fixed & Dam. & Rem. \\
\midrule
Corrupted & 30.0 & 0 & 0 & 0 & 29.7 & 0 & 0 & 0 & 30.0 & 0 & 0 & 0 \\
Loss filter & 16.9 & 0 & 0 & 126 & 19.9 & 0 & 0 & 115 & 18.4 & 0 & 0 & 183 \\
Confident learning & 15.3 & 0 & 0 & 200 & 21.7 & 0 & 0 & 28 & 15.7 & 0 & 0 & 35 \\
GPT-Agree & 8.9 & 0 & 0 & 164 & 5.9 & 0 & 0 & 35 & 6.7 & 0 & 0 & 28 \\
GPT-Judge & 8.1 & 714 & 77 & 280 & 4.4 & 431 & 13 & 30 & 4.4 & 869 & 21 & 24 \\
LLM-AutoDP & 3.0 & 0 & 0 & 305 & 0.2 & 0 & 0 & 27 & 16.7 & 519 & 1 & 0 \\
\midrule
\cado{} & 7.6 & 508 & 6 & 154 & 11.8 & 253 & 0 & 2 & 0.0 & 1,166 & 0 & 0 \\
\bottomrule
\end{tabular}
\end{table*}

\begin{figure*}[t]
\centering
\includegraphics[width=\textwidth]{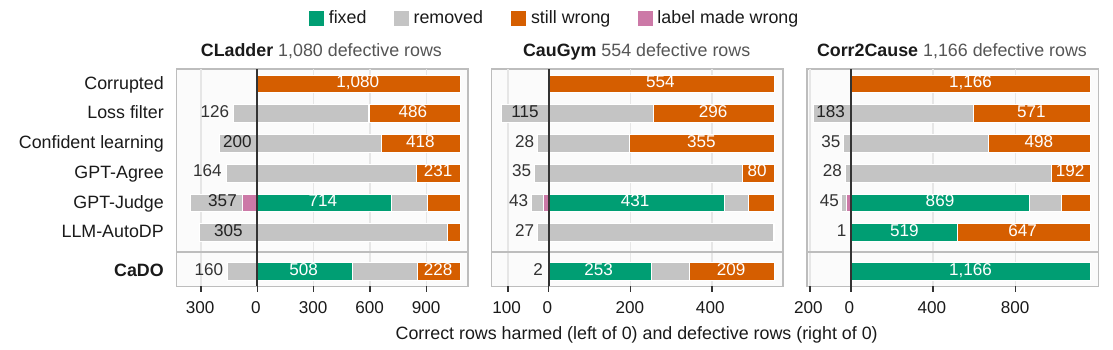}
\caption{What each method does to the corrupted training data, scored against the hidden ground truth as in Table~\ref{tab:dq}. Right of 0: the defective rows, which a method fixes (keeps with a corrected label), removes or leaves wrong. Every method receives the same defective rows. Left of 0: the correct rows that it harms by changing their label to a wrong one or by removing them, with their total. Rows that a method adds are not shown. On CLadder, GPT-Judge changes 77 correct labels to wrong ones and \cado{} 6.}
\label{fig:dq}
\end{figure*}

\begin{table}[t]
\centering
\caption{Verification of the corrupted training sets against the hidden ground truth. Recall: share of the rows with each example-level defect that the Diagnoser flags. Last row: repairs applied by \cado{} (high tier) and, in brackets, how many are wrong.}
\label{tab:verify}
\footnotesize
\setlength{\tabcolsep}{4pt}
\begin{tabular}{lrrr}
\toprule
& CLadder & CauGym & Corr2Cause \\
\midrule
Flag precision (\%) & 95.0 & 99.8 & 100.0 \\
Flag recall (\%), all defects & 74.0 & 90.8 & 100.0 \\
\quad label flipped & 80.6 & 90.9 & 100.0 \\
\quad number changed & 81.9 & 92.9 & 100.0 \\
\quad link changed & 50.5 & 90.4 & 100.0 \\
\quad question changed & 85.7 & 92.1 & 100.0 \\
\quad weak-teacher solution & 71.3 & 87.5 & 100.0 \\
Clean rows flagged (\%) & 1.0 & 0.1 & 0.0 \\
Repairs applied (wrong) & 514 (6) & 253 (0) & 1,166 (0) \\
\bottomrule
\end{tabular}
\end{table}

\begin{table}[t]
\centering
\caption{Dataset-level defects on CLadder (Qwen3-4B, test). Left: accuracy (\%) on the three under-covered query types (coverage gap). Right: on the three label-skewed types, the excess share of ``yes'' predictions over the true share (points) and the accuracy on their ``no'' items.}
\label{tab:types}
\footnotesize
\setlength{\tabcolsep}{3pt}
\begin{tabular}{lrrrrr}
\toprule
& \multicolumn{3}{c}{Under-covered} & \multicolumn{2}{c}{Label-skewed} \\
\cmidrule(lr){2-4}\cmidrule(lr){5-6}
Training data & ETT & NIE & ATE & Yes excess & ``No'' acc. \\
\midrule
Corrupted & 29.3 & 82.3 & 93.4 & +24.8 & 49.1 \\
Loss filter & 48.8 & 83.7 & 96.2 & +24.5 & 49.5 \\
Confident learning & 40.8 & 82.0 & 93.8 & +26.5 & 45.4 \\
GPT-Agree & 48.3 & 75.3 & 93.4 & +4.2 & 83.2 \\
GPT-Judge & 64.2 & 66.2 & 89.7 & +4.0 & 83.6 \\
LLM-AutoDP & 81.1 & 68.8 & 91.8 & +2.8 & 90.2 \\
\midrule
\cado{} & 91.8 & 84.4 & 93.8 & +0.8 & 96.7 \\
\quad without generation & 53.5 & 81.5 & 94.0 & +1.7 & 94.8 \\
Clean (oracle) & 97.0 & 96.6 & 99.0 & +0.0 & 96.3 \\
\bottomrule
\end{tabular}
\end{table}

\subsection{Quality of the Repaired Data}
\label{sec:dq}

Scoring each method's training data against the hidden ground truth, in Table~\ref{tab:dq} and Fig.~\ref{fig:dq}, explains the accuracy results, and Table~\ref{tab:verify} measures the Diagnoser directly.

\textbf{Verification is precise.} More than 95\% of the examples that the Diagnoser flags are defective, and at most 1\% of the clean rows are flagged. Of the repairs that \cado{} applies, 6 of 514 are wrong on CLadder and none on CauGym and Corr2Cause. The reading itself is rarely the problem, since with the Formalizer Agent's greedy reading the engine reaches the correct answer for 98.2\% of CLadder's and 99.4\% of CauGym's dev questions. Recall is lowest for changed causal links on CLadder, at 50.5\%. When a stated link is removed or reversed, the readings often disagree or the query becomes unidentifiable, so the Diagnoser abstains and Filter removes the row. Of CLadder's 1,080 defective rows, \cado{} fixes 508, removes 344 and keeps 228 with a wrong label.

\textbf{LLM judges damage correct data.} GPT-Judge corrects many labels, 714 on CLadder, but it also changes 77 correct CLadder labels to wrong ones, 13 on CauGym and 21 on Corr2Cause. Each of these is a question that the judge itself answered wrongly, the circularity of Fig.~\ref{fig:teaser}(a). \cado{} changes 6, 0 and 0, because it computes the answer and acts only on firm ground, so compared with GPT-Judge, it damages 92.2\% fewer correct labels on CLadder.

\textbf{Filtering removes correct data and leaves the dataset-level defects.} The filtering baselines remove defective and correct rows alike. LLM-AutoDP removes 305 correct CLadder rows to reach 3.0\% wrong labels, and GPT-Agree removes 164. Removal cannot add missing query types or labels, and none of the baselines repairs the coverage gap or the label skew, as Table~\ref{tab:types} shows. On the label-skewed CLadder types, an external evaluation model trained on corrupted data overpredicts ``yes'' by 24.8 points, and only \cado{} brings this excess below one point. On the under-covered questions about the effect of treatment on the treated, accuracy rises from 29.3\% with corrupted data to 81.1\% after LLM-AutoDP and 91.8\% after \cado{}, against 97.0\% with clean data.

\textbf{Repair and generation complement each other.} \cado{} trains better models than LLM-AutoDP on CLadder, although it keeps more wrong labels, 7.6\% against 3.0\%. It applies only high-tier repairs and abstains elsewhere, and it adds 1,382 engine-labelled examples where the data is thin. On Corr2Cause, where the reading is exact, \cado{} fixes all 1,166 defective rows without damaging any.

\begin{table}[t]
\centering
\caption{Components of \cado{} (Qwen3-4B, headline test): variant minus \cado{} in points, at equal seeds (3, or 8 where the sequential rule added seeds). $^{\dagger}$: 95\% interval excludes 0. =: the variant produces \cado{}'s training set. Generated rows: those of \cado{}. --: not run.}
\label{tab:ablation}
\footnotesize
\setlength{\tabcolsep}{2.5pt}
\begin{tabular}{lrrr}
\toprule
Variant & CLadder & CauGym & Corr2Cause \\
\midrule
\multicolumn{4}{l}{\emph{Verification}} \\
\quad GPT-Judge instead of the engine & $-$3.3$^{\dagger}$ & $-$0.4 & $-$0.4 \\
\quad one Formalizer Agent reading & +0.6 & +2.1$^{\dagger}$ & = \\
\quad no Critic Agents & +0.5 & +3.2$^{\dagger}$ & = \\
\quad repairs of every tier & +0.8 & +2.5$^{\dagger}$ & = \\
\multicolumn{4}{l}{\emph{Generation}} \\
\quad no generation & $-$4.9$^{\dagger}$ & $-$0.8 & $-$1.9$^{\dagger}$ \\
\quad LLM labels for the generated rows & $-$6.0$^{\dagger}$ & $-$2.4$^{\dagger}$ & $-$2.5$^{\dagger}$ \\
\multicolumn{4}{l}{\emph{Data volume}} \\
\quad corrupted data + generated rows & $-$2.9$^{\dagger}$ & -- & $-$2.0$^{\dagger}$ \\
\quad generated rows only & $-$2.6$^{\dagger}$ & -- & -- \\
\bottomrule
\end{tabular}
\end{table}

\subsection{Component Analysis and Data-Volume Controls}
\label{sec:ablation}

Changing one component at a time and controlling for the added rows shows where \cado{}'s gains come from, as Table~\ref{tab:ablation} reports for Qwen3-4B.

\textbf{The engine.} Replacing the engine's verdicts with GPT-Judge's decisions, with everything else unchanged, costs 3.3 points on CLadder. On CauGym and Corr2Cause the difference in accuracy is below one point, but the data is clearly worse, with 13 and 21 correct rows damaged and 65 and 143 defective rows left wrong. On CauGym, with eight seeds per arm, the dev split shows a significant loss of 2.6 points.

\textbf{Generation.} Generation helps because the engine labels it. Without Generate and Rebalance, accuracy drops by 4.9 points on CLadder and 1.9 on Corr2Cause. When we keep the same generated questions but let the LLM write their labels and solutions, accuracy drops by 6.0, 2.4 and 2.5 points, although the LLM's labels differ from the engine's on only 2.8\%, 1.1\% and 2.2\% of the generated rows.

\textbf{Critic Agents and the tier rule.} The extra readings of the Formalizer Agent, the Critic Agents and the high-tier rule trade recall for safety. On CLadder they cut wrong repairs from 63 of 772, or 8.2\% in a variant with one reading, no Critic Agents and no tiers, to 6 of 514, or 1.2\%, and without the Critic Agents 36 correct rows are damaged instead of 6. Accuracy shows no detectable difference. On CauGym, where the Formalizer Agent is nearly perfect, the same caution holds back 206 correct repairs and costs 2.5 points against repairs of every tier and 3.2 points against no Critic Agents. Our pre-registered default is the cautious one, but a dataset with an accurate Formalizer Agent can afford to apply every tier. On CauGym, the gain of a single reading, +2.1, shrinks to +1.1 against all eight seeds of \cado{}.

\textbf{Smaller operators.} The remaining operators matter little on CLadder, where removing RationaleReplace, Debias or Rebalance, generating towards a uniform mix of types, or keeping the unresolved rows changes test accuracy by less than one point. The planted cue turned out to be inert, because the stated numbers determine the answer. Inserted into 400 clean test questions, it changed no model's accuracy by more than 0.7 points. Debias removes it from all 638 training rows that carry it, without effect.

\textbf{Data volume.} \cado{}'s lead does not come from its larger output. It outputs 25\% to 43\% more rows than it receives, since it generates rows to fill gaps and balance labels, and three controls separate the effect of volume from that of verification. First, \cado{}'s output, subsampled to the size of a baseline's output, still beats it, by +3.7 \ci{+1.1}{+6.7} over LLM-AutoDP and +8.3 \ci{+6.1}{+10.4} over GPT-Judge on CLadder, and by +3.2 \ci{+1.0}{+5.2} over LLM-AutoDP on Corr2Cause. On CauGym, subsampled to GPT-Agree's output, it is ahead by +1.4 \ci{-0.1}{+3.0} with eight seeds per arm, which is inconclusive. Second, adding \cado{}'s generated rows to the corrupted data isolates the value of repair, and \cado{} stays ahead by 2.9 points on CLadder and 2.0 on Corr2Cause. Third, training only on 4,484 engine-labelled generated rows, as in engine-based data generation~\cite{dai2026unico}, is 2.6 points below \cado{} on CLadder. Repairing the given data beats discarding it.

\begin{figure*}[t]
\centering
\includegraphics[width=\textwidth]{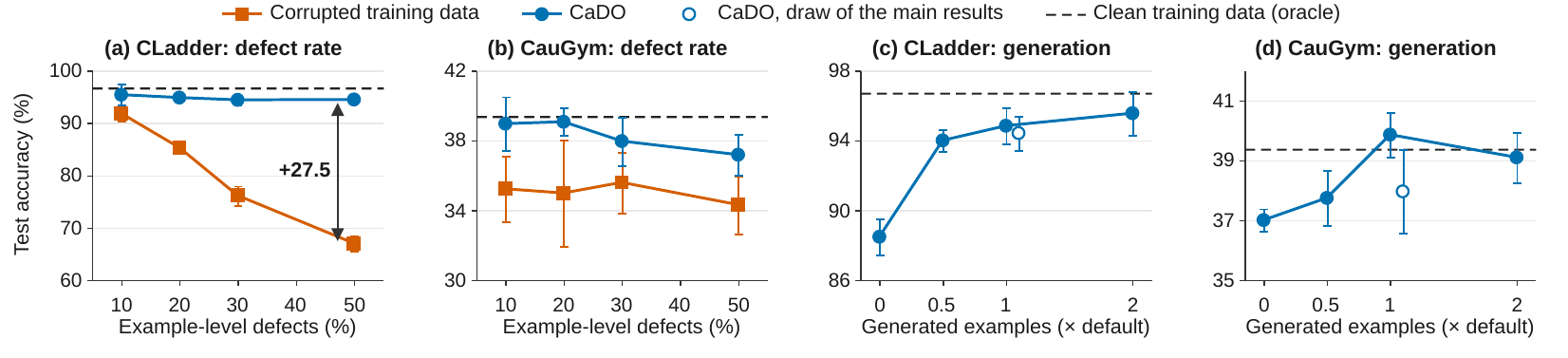}
\caption{Sensitivity of \cado{} (Qwen3-4B, headline test sets). (a,~b)~Share of example-level defects, with the dataset-level defects unchanged. (c,~d)~Amount of generated data as a multiple of \cado{}'s default, drawn from an enlarged pool. The hollow marker is the draw used in the main results. Mean over seeds with bars of $\pm$1 standard deviation. Dashed: clean training data. The label in (a) is the paired difference at 50\% defects.}
\label{fig:sweeps}
\end{figure*}

\subsection{Exact Verification versus Fine-Tuning Feedback}
\label{sec:planner}

Choosing the plan by fine-tuning feedback is never significantly better than \cado{}'s fixed rules. \search{} beats the strongest baseline significantly in 14 of 18 settings, while \cado{} beats \search{} in 3. Against Random search with the same operators, budget and proxy, \search{} is significantly better in 3 settings, worse in 1 and not significantly different in 14, whereas \cado{}, which searches nothing, is better than Random search in 7 settings and never worse. The Planner Agent does use the diagnosis, since one that sees only the scores loses 3.5 points on CLadder by never choosing Repair. A Reflector Agent that writes lessons and enforces the noise margin adds nothing detectable.

The reason is measurement noise. On CLadder with Qwen3-4B, the standard deviation of a single fine-tuning run is 1.2 to 1.4 points, and 2.9 points for the half-data proxy runs of the search, while all reasonable pipelines lie within about one point of each other. Among the 34 arms with a test accuracy of at least 93\%, the correlation between dev and test means is only 0.29. A search scored by proxy fine-tuning cannot resolve the differences between good pipelines, and picking the best of 20 noisy scores favours lucky draws. Exact verification has no such floor, because \cado{} checks each decision per example before any model is trained.

\subsection{Sensitivity to the Defect Rate and the Generation Amount}
\label{sec:sweeps}

\cado{}'s accuracy stays nearly flat as example-level defects grow, while the corrupted data degrades. Fig.~\ref{fig:sweeps}(a,~b) varies their share from 10\% to 50\% over nested subsets of the defective rows. On CLadder, the corrupted data falls from 91.8\% to 67.0\%, while \cado{} stays between 94.4\% and 95.4\%, so its gain grows from 3.6 to 27.5 points. On CauGym, \cado{} is significantly ahead of the corrupted data at every rate, by 2.4 to 4.1 points. Each row is checked on its own, so more defects only mean more repairs.

More generated data helps on CLadder and saturates on CauGym. Fig.~\ref{fig:sweeps}(c,~d) scales the amount of generated data. On CLadder, accuracy rises monotonically from 88.5\% without generation to 95.6\% at twice the default amount. On CauGym, whose metric depends mostly on arithmetic, it rises from 37.0\% to 39.9\% at the default amount and no further. There, a fresh draw of the default amount scores 1.9 points above the draw of the main results, the hollow marker, so the main results do not rest on a favourable draw.

\subsection{Results on the Original Datasets}
\label{sec:real}

A repair method must agree with labels that are already right, so we first ran the engine on the datasets' hidden metadata instead of the Formalizer Agent's reading. Of CLadder's 10,112 labels, the engine agrees with 9,265 beyond the rounding of the stated numbers, leaves 825 undetermined within that rounding and cannot compute 9, whose stated numbers make a ratio undefined. It disagrees with the remaining 13, of which 6 are genuine contradictions in CLadder and 7 follow from its two conventions. The engine reproduces all 2,500 CauGym effect labels exactly, 999 of the 1,000 probability bounds within rounding, and 97.1\% of Corr2Cause's dev and test labels. The remaining Corr2Cause labels are not entailed by their own premises.

\cado{}, run unchanged on the original training sets, leaves model accuracy unchanged. For Qwen3-4B and Qwen3-8B, the six differences from training on the data as is lie between $-0.9$ and $+0.5$ points, and none is significant. On the original CLadder, \cado{} changed the labels of 6 of the 6,963 rows that our clean copy confirms. On the original Corr2Cause, the engine contradicts 206 official labels, 3.4\%, whose premises do not entail them, and \cado{} changes them. Models trained on its output stop reproducing such labels, 5\% against 65\% of the 37 dev items concerned, and on the test items whose labels the engine confirms, the 8B model improves by +1.2 \ci{+0.5}{+2.0}. We found one real cost. \cado{}'s generated rows raise the share of ``valid'' labels from 15.3\% to 19.9\%, and the 4B model loses 2.2 \ci{-3.4}{-1.0} points on the engine-confirmed items of the refactorized test set because it overpredicts ``valid''.

\begin{table}[t]
\centering
\caption{Processing cost per corrupted training set: LLM calls (millions of prompt and completion tokens) and GPU-hours of pipeline search. \cado{}'s calls are the Critic Agents' checks of the flagged rows.}
\label{tab:cost}
\footnotesize
\setlength{\tabcolsep}{3pt}
\begin{tabular}{lrrr}
\toprule
& CLadder & CauGym & Corr2Cause \\
\midrule
\multicolumn{4}{l}{\emph{LLM calls (M tokens)}} \\
\quad \cado{} & 1,635 (2.3) & 967 (2.0) & 2,334 (3.4) \\
\quad GPT-Agree & 3,599 (1.8) & 1,863 (1.7) & 3,886 (5.3) \\
\quad GPT-Judge & 3,599 (3.0) & 1,863 (2.6) & 3,886 (5.3) \\
\quad LLM-AutoDP & 6,192 (5.2) & 3,595 (6.0) & 6,629 (9.0) \\
\midrule
\multicolumn{4}{l}{\emph{GPU-hours of pipeline search}} \\
\quad \cado{} & 0 & 0 & 0 \\
\quad \search{} & 13.9 & 19.7 & 24.5 \\
\quad LLM-AutoDP & 10.9 & 14.1 & 22.4 \\
\bottomrule
\end{tabular}
\end{table}

\subsection{Cost}
\label{sec:cost}

\cado{} needs fewer LLM calls than the per-row LLM baselines and no pipeline search, as Table~\ref{tab:cost} shows. Because \cado{} sends to the Critic Agents only the rows that the Diagnoser would change or remove, it needs 45\% to 60\% of the LLM calls of GPT-Judge or GPT-Agree, which call it for every row, and a quarter to a third of LLM-AutoDP's. Its token use is close to that of the per-row baselines. Its other costs are local and paid once per dataset, about two GPU-hours to train the Formalizer Agent for CauGym, about ten minutes on one GPU to read every CLadder row three times, and 188 LLM calls for CLadder's generation pool, none for Corr2Cause.

\subsection{Summary of Findings}
\label{sec:summary}

Based on the experimental results above, we draw four conclusions. First, based on the main comparison, we conclude that \cado{}'s output trains more accurate external evaluation models than that of the cleaning baselines. Second, based on the quality of the repaired data, we conclude that executable validation makes repairs precise. \cado{} abstains when a reading is uncertain, whereas an LLM judge damages correct labels. Third, based on the component analysis, the data-volume controls and the comparison with \search{}, we conclude that the gains come from the engine's labels rather than from more data, and that choosing plans by fine-tuning feedback brings no further gain. Finally, based on the original datasets, we conclude that \cado{} also repairs real defects without harming correct data. It rarely changes a correct label and leaves model accuracy unchanged, yet it corrects labels that their premises do not entail. Taken together, these findings show that computing correctness instead of judging it makes \cado{}'s repairs both precise and useful for training.

\section{Related Work}
\label{sec:related}

This section relates \cado{} to three lines of work. Unlike them, \cado{} runs a causal engine backward over existing training data, so that exact computation decides what to repair.

\textbf{Causal datasets and engines.} CLadder~\cite{jin2023cladder}, Corr2Cause~\cite{jin2024corr2cause} and CauGym~\cite{chen2026caugym} are generated forward from sampled causal models, related benchmarks include CaLM~\cite{chen2024calm} and CounterBench~\cite{chen2026counterbench}, and UniCo~\cite{dai2026unico} and training with verifiable rewards~\cite{lu2025rlvr} generate new training data with exact inference. Audits still find incorrect, under-specified and shortcut-prone items in such data~\cite{dai2026unico,li2025cdcr,clarke2026procedural,yu2026caliper,yang2024critical}, and CausalBench~\cite{wang2024causalbench} checks GPT-generated answers with an engine and resolves disagreements by hand. Translate-then-execute methods let a solver answer a problem that an LLM translated into a program or a formal specification~\cite{gao2023pal,chen2023pot,lyu2023faithfulcot,ye2023satlm,pan2023logiclm,olausson2023linc}. In causality, agents call causal libraries~\cite{jiang2024llm4causal,wang2025causalcopilot}, and DoVerifier~\cite{he2026doverifier} checks model-written do-calculus expressions. In \cado{}, the engine runs offline, and a disagreement with a stored label reveals a defect.

\textbf{Data cleaning and preparation.} Data management research repairs errors with constraints~\cite{chu2013holistic} and probabilistic inference~\cite{rekatsinas2017holoclean}, detects them with few labels~\cite{mahdavi2019raha,mahdavi2020baran}, and cleans data for downstream models~\cite{krishnan2016activeclean,krishnan2017boostclean,li2021cleanml,wu2020complaint,chu2016datacleaning,whang2023datacollection}. \cado{} shares the structure of detection, diagnosis and repair, but its integrity constraint is semantic and checked by execution, since an example must have the answer that the engine derives from its premises. Solvers also help experts relabel logic~\cite{brunello2026fixing,ni2026agentified} and text-to-SQL data~\cite{jin2026texttosql,zhu2026birdplatinum}, whereas \cado{}'s engine is an automatic oracle. LLMs clean tables~\cite{narayan2022wrangle,zhang2024jellyfish,li2024tablegpt,ni2025zeroed,yan2024gidcl,li2025autodcworkflow}, and Data-Juicer~\cite{chen2024datajuicer}, Star-Agents~\cite{zhou2024staragents}, SQL-Factory~\cite{li2025sqlfactory} and LLM-AutoDP~\cite{huang2026llmautodp} process or generate LLM training data and validate it with an LLM's judgement, a heuristic, a downstream score or, in SQL-Factory, execution, which \cado{} brings to causal text.

\textbf{Label noise and verified synthetic data.} Confident learning~\cite{northcutt2021confident}, small-loss selection~\cite{han2018coteaching} and data selection for LLMs~\cite{chen2024alpagasus,liu2024what,li2024quantity,zhou2023lima} drop examples by model statistics~\cite{song2023learning}, losses or LLM ratings, although LLM judges are biased~\cite{zheng2023judging,wang2024large} and label errors occur even in test sets~\cite{northcutt2021pervasive,gema2025mmlu}. These methods filter rather than repair. Unit tests, database execution and proof assistants filter new code, SQL and mathematics data~\cite{wei2024selfcodealign,xu2025kodcode,li2025omnisql,xin2024deepseekprover,ying2024lean,lu2024mathgenie}, self-training keeps rationales that reach the gold answer~\cite{zelikman2022star,yuan2023scaling,singh2024beyond}, and axiomatic training~\cite{vashishtha2025teaching} generates causal demonstrations. These pipelines create new data or trust gold labels, whereas \cado{} repairs existing data and questions its labels.

\section{Conclusion}
\label{sec:conclusion}

LLMs learn causal reasoning largely from generated datasets that inevitably contain noise and errors, which can propagate into downstream models. In this paper, we proposed \cado{}, an automated multi-agent framework for causal dataset diagnosis, repair, and generation. The Formalizer Agent translates natural language questions into formal causal queries for reliable answer verification. A dual-level diagnosis mechanism employs independent Critic Agents to identify sample-level errors and analyzes query-type coverage and label distributions to detect dataset-level imbalances. Based on these diagnoses, reusable repair operators correct erroneous samples using formal causal semantics, while Explorer, Expander, and Formalizer Agents collaboratively generate diverse and semantically consistent samples to improve dataset balance. Experiments on three datasets demonstrated that \cado{} achieves precise error detection, damages far fewer correct labels than an LLM judge, and produces datasets that train more accurate models than the strongest of five cleaning baselines. Our work provides a reliable approach to improving causal reasoning datasets without extensive manual intervention. Future work will extend the framework to more causal query types and naturally occurring defects.

\bibliographystyle{IEEEtran}
\bibliography{refs}

\end{document}